\documentclass[runningheads,a4paper]{llncs}

\usepackage{amssymb}
\usepackage{graphicx}
\usepackage{amsmath}\usepackage{amsmath}

\usepackage{subfigure}

\usepackage{url}
\urldef{\mailsa}\path|{devinder.kumar, agchung, mjshafie, alexander.wong}@uwaterloo.ca|
\urldef{\mailsc}\path|farzad.khalvati@sri.utoronto.ca, masoom.Haider@sunnybrook.ca|    
\newcommand{\keywords}[1]{\par\addvspace\baselineskip
\noindent\keywordname\enspace\ignorespaces#1}

\begin{document}

\mainmatter  

\title{Discovery Radiomics for Pathologically-Proven Computed Tomography Lung Cancer Prediction}

\titlerunning{Discovery Radiomics for Pathologically-Proven Lung Cancer Prediction}

%
%
\author{Devinder Kumar\textsuperscript{1*}%
\and Audrey G. Chung\textsuperscript{1}\and Mohammad J. Shaifee\textsuperscript{1}\and Farzad Khalvati\textsuperscript{2,3}\and
Masoom A. Haider\textsuperscript{2,3}\and Alexander Wong\textsuperscript{1}}
\authorrunning{Kumar et. al.}

\institute{\textsuperscript{1}Department of Systems Design Engineering, University of Waterloo, ON, Canada \\
\textsuperscript{2}Department of Medical Imaging, University of Toronto, ON, Canada\\\textsuperscript{3}Sunnybrook Health Sciences Centre, Toronto, Ontario, Canada, M4N 3M5
\mailsa\\
\mailsc\\
}

%
%

\toctitle{Discovery Radiomics Kumar et. al.}
\tocauthor{Kumar et. al.}
\maketitle

\vspace{-0.4cm}
\begin{abstract}
Lung cancer is the leading cause for cancer related deaths. As such, there is an urgent need for a streamlined process that can allow radiologists to provide diagnosis with greater efficiency and accuracy.  A powerful tool to do this is radiomics: a high-dimension imaging feature set. In this study, we take the idea of radiomics one step further by introducing the concept of \textit{discovery radiomics} for lung cancer prediction using CT imaging data. In this study, we realize these custom radiomic sequencers as deep convolutional sequencers using a deep convolutional neural network learning architecture. To illustrate the prognostic power and effectiveness of the radiomic sequences produced by the discovered sequencer, we perform cancer prediction between malignant and benign lesions from 97 patients using the pathologically-proven diagnostic data from the LIDC-IDRI dataset. Using the clinically provided pathologically-proven data as ground truth, the proposed framework provided an average accuracy of 77.52\% via 10-fold cross-validation with a sensitivity of 79.06\% and specificity of 76.11\%, surpassing the state-of-the art method.
\keywords{Radiomics, discovery radiomics, deep convolutional neural network, lung nodules}
\end{abstract}

\vspace{-1cm}
\section{Introduction}
\vspace{-0.2cm}
According to the American Cancer Society~\cite{acs2015}, lung cancer is the second most diagnosed form of cancer in the United States, second only to prostate cancer in males and breast cancer in females. Furthermore, lung cancer remains the leading cause of cancer-related deaths in the United States, accounting for approximately 27\% of all cancer-related deaths.  A similar report by the Canadian Cancer Society~\cite{ccs2015} estimates the number of new cases of lung cancer to be the highest of all cancers in Canada, and it is also the leading cause of cancer-related deaths in Canada, accounting for approximately 26\% of all cancer-related deaths.  Early screening and diagnosis of lung cancer at a more treatable stage of the disease can play a pivotal role in improving the survival rates for such patients.

One of the biggest emerging areas in recent years related to quantitative cancer screening and diagnosis is radiomics~\cite{lambin2012radiomics,balagurunathan2015radiomic}, which involves the high-throughput extraction and analysis of a large number of imaging-based features for quantitative characterization and analysis of tumour phenotype.  The use of radiomics-driven approaches allow for a more objective and quantitative evaluation and diagnosis of cancer, which can significantly reduce inter-observer and intra-observer variability and improve diagnostic accuracy and efficiency compared to current qualitative cancer assessment strategies.  In a comprehensive study by Aerts et. al.~\cite{aerts2014decoding} involving more than 1000 patients across seven datasets, it was shown that radiomics can be used to obtain phenotype differences between tumours that can have clinical significance and prognostic value. The idea behind the success of radiomics as revealed by radio-genomics is the radiomic sequences being generated were able to capture quantitative features that define intra-tumour heterogeneity, which is directly related to the underlying gene-expression patterns~\cite{aerts2014decoding}. Other radiomics-driven approaches have been investigated for lung cancer prediction~\cite{anirudh2016lung,shen2015multi,shen2017multi}. For example, \mbox{Anirudh et. al.}~\cite{anirudh2016lung} used automated weakly labeled data with 3D CNNs to classify lung nodules for SPIE-LUNGx challenge.  In~\cite{orozco2015automated}, \mbox{Orozco et. al.}, used a radiomic sequence consisting of wavelet features and illustrated its effectiveness on a smaller sub-set of CT images from the LIDC-IDRI dataset. Shen et. al.~\cite{shen2015multi} introduced a convolutional radiomic sequencer, based on multi-crop windows around the lung nodules effectively and later extends the framework~\cite{shen2017multi} to rate the suspiciousness of nodules. One important thing to note here is that, all of the above mentioned literature explored the radiologist driven nodule ratings for predicting the malignant of nodules.

In this study, we use the concept of \textit{discovery radiomics} for lung cancer prediction, where we discover custom radiomic sequencers tailored for lung cancer characterization and prediction using pathologically-proven diagnostic dataset. To realize the concept of discovery radiomics for lung cancer prediction in this study, we introduce a deep convolutional radiomic sequencer that is discovered using a deep convolutional neural network architecture based on CT imaging data and pathologically-proven diagnostic data from past patients in a medical imaging archive. This methodology of using a more challenging pathologically-proven diagnostic data as ground truth distinguishes this study from the above mentioned literature that are based only on radiologist provide malignancy scores. In recent past, only two studies were proposed ~\cite{shen2016learning,kumar2015lung}, which also explores the pathologically-proven dataset from LIDC-IDRI dataset for predicting lung nodule malignancy. In~\cite{kumar2015lung}, the authors proposed an autoencoder based unsupervised approach for predicting malignancy of the lung nodules. More recently, Shen et. al~\cite{shen2016learning} proposed a method using CNN incorporated with multiple instance learning for classifying the pathologically-proven diagnostic data. In~\cite{shen2016learning}, the authors performed knowledge (weight) transfer from CNN trained on large non-diagnostic data to a different CNN for the task for diagnostic malignancy prediction.  The authors compared different methodologies and prove that their method is the current state-of-the art in this domain. In this study, we surpass the state-of-the art results stated in~\cite{shen2016learning} which proves the effectiveness for the proposed discovery radiomics approach.

The main contributions of this study can be summarized as follows: (i) An effective malignancy prediction framework can be developed with limited pathologically proven diagnostic data without any external knowledge or large radiologist based annotated dataset. (ii) We provide experimental evidence that the proposed discovery radiomics framework is more effective in predicting pathologically-proven malignancy than the state-of-the art method~\cite{shen2016learning}.   

\begin{figure}[t]
\centering
  \includegraphics[clip,trim = 0cm 3cm 0cm 0cm,height = 6.0 cm,width=0.8\textwidth]{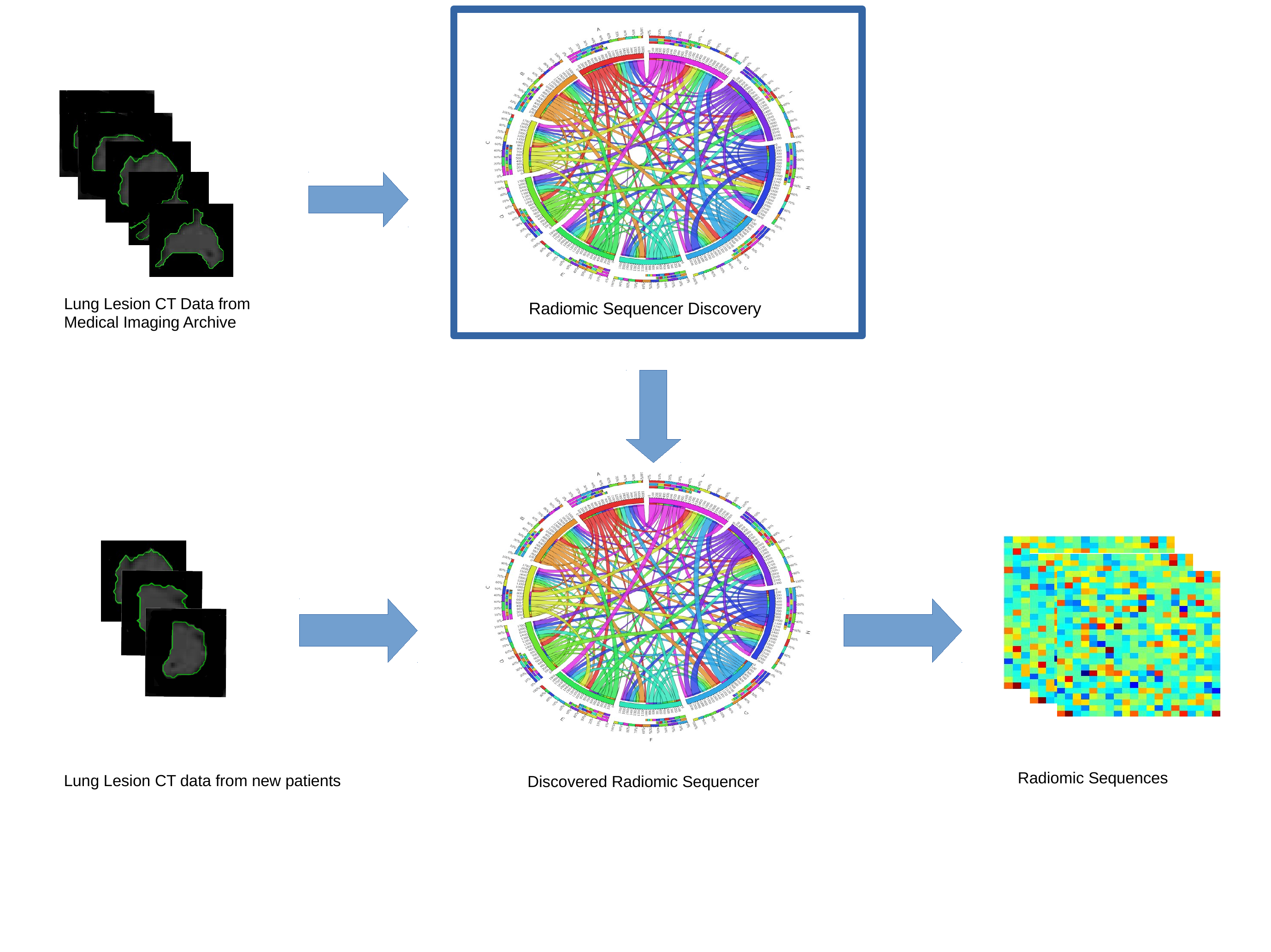}
  \caption{Overview of the proposed discovery radiomics framework for CT lung cancer prediction. A custom radiomic sequencer is discovered directly using the wealth of lung CT data, hand annotations by the radiologist, and diagnostic results available from the imaging archive.  The discovered radiomic sequencer can then be used to generate custom radiomic sequences for new patients using their lung CT data.}
  \label{rad}
  \vspace{-0.6cm}
\end{figure}
\vspace{-0.5cm}
\section{Methodology}
\vspace{-0.2cm}
This section presents the methodology used in this study in detail. The various steps involved in the proposed discovery radiomics framework for lung cancer prediction is shown in the Fig.~\ref{rad}. First, using the wealth of CT imaging data, radiologist annotations, and pathologically-proven diagnostic results for past patients available in a medical imaging archive, a custom radiomic sequencer is discovered for generating radiomic sequences composed of abstract, quantitative imaging-based features that characterize tumour phenotype.  Once we obtain the discovered radiomic sequencer, we can apply this sequencer to any new patient case to extract radiomic sequences tailored for lung cancer characterization and prediction.
\vspace{-0.4cm}
\subsection{Patient Data Collection}
\vspace{-0.2cm}
For learning the custom radiomic sequencer in the proposed discovery radiomics framework, a set of lung CT images, radiologist annotations, and pathologically-proven diagnostic results from past patients is needed.  In this study, we use a subset of LIDC-IDRI~\cite{armato2011lung} dataset. Seven different academic and eight medical imaging companies collaborated to build the LIDC-IDRI database, which consists of data for 1,010 patients, each of which contains clinical helical thoracic CT scans and associated data of two-phase annotation process performed by four experienced thoracic radiologists. For annotating the nodules in the lung CT scans, a two phase annotation process was finalized for obtaining the interpretation from up to four different radiologists for a single CT scan. The radiologists marked each identified lesion as either a nodule of size $>=$ 3mm, $<$ 3mm or non-nodule $>=$ 3mm. The database contains a total of 7,371 lesions marked as nodule by at least one radiologists with 2,699 of them marked $>=$ 3mm. We selected a subset of the LIDC-IDRI of 97 patient cases, for which definite pathologically-proven diagnostic results were available. The diagnostic results were obtained at two levels: i) patient level and ii) nodule level. At each level, the lesions in lung were marked as either: 0 - Unknown (no data),1 - Benign, 2 - Primary malignancy or 3 - Metastatic lesion with extra-thoracic primary malignancy.

These ratings were obtained by using either biopsy, surgical resection, progression or review of radiological images to show two years of stable nodule size. In this study, we take the lesions marked $>=$ 3mm for only the cases with ratings of 1, 2 or 3 and classifying the ratings 2 and 3 as malignant and rating 1 as benign. This resulted in 69 patient cases for malignant and 28 patient cases for benign lesions, with a total of 608 benign lesions and 3,698 malignant lesions. It is important to note here that lesions $>=$ 3mm were considered to facilitate for a fair comparison with other state-of-the art methods~\cite{shen2016learning,kumar2015lung}.
\vspace{-0.4cm}
\subsection{Data Augmentation}
\vspace{-0.2cm}
For building an effective radiomics sequencer that can well represent and characterize different types of the lung lesions and have a prognosis efficacy similar to the pathologically-proven diagnosis level results, there is a need of larger datasets for the radiomic sequencer discovery process. Therefore, to further enrich our dataset, we take each lesion candidate associated with pathologically-proven diagnostic dataset and perform spatial deformation using rotations. Using this, we ended with a total of 42,340 lesion candidates to be used for radiomic sequencer discovery.  More specifically, each lesion for the malignant cases is rotated by 45 degrees in total eight different variations for the particular lesion to get 29,956 malignant lesions. As the initial number of cases marked with benign were very less and it is ideal to have a balanced dataset for the radiomic sequencer discovery process, we rotated each benign lesion by 10 degrees from 0 to 360 degrees resulting in 22,384 benign lesions for creating near equal lesion candidate ratio for both the categories.

\begin{figure}[t]
\centering
  \vspace{-0.2cm}
  \includegraphics[clip,trim = 0cm 8cm 0cm 6cm, width=0.8\textwidth]{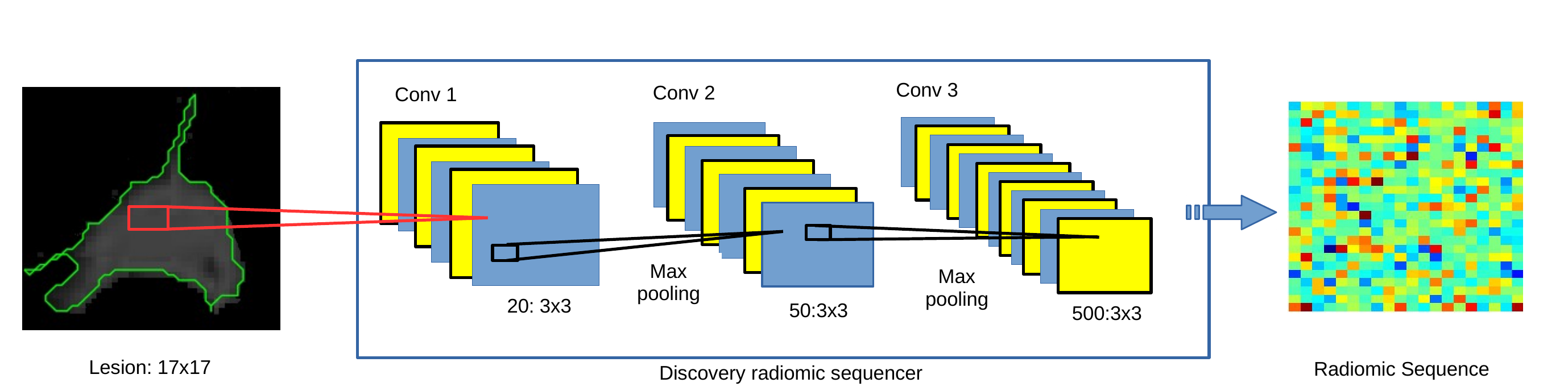}
  \caption{Architecture of the deep convolutional radiomic sequencer consists of 3 convolutional sequencing layers and 2 max pooling sequencing layers, which outputs a radiomic sequence with 500 abstract imaging-based features.}
  \label{cnn_out}
  \vspace{-0.6cm}
\end{figure}

\begin{table}[ht]
	\centering
	\caption{Summary of number of receptive fields and receptive field size at each convolutional sequencing layer.}
	\vspace{-0.3cm}
	\begin{tabular}{|c|c|c|}
		\hline
		\textbf{Sequencing Layer} & \textbf{Number of Receptive Fields} & \textbf{Receptive Field Size} \\
		\hline
		1							& 	$20$  		&	$3 \times 3$	\\
		2							& 	$50$  		&	$3 \times 3$	\\
		3							& 	$500$  		&	$3 \times 3$	\\
		\hline
	\end{tabular}
	\vspace{-0.3cm}
	\label{tab_Kernels}
\end{table}	
\vspace{-0.4cm}
\subsection{Radiomic Sequencer}
\vspace{-0.2cm}
Given the CT imaging data, radiologist annotations, and pathologically-proven diagnostic results for past patients, the next step is to discover a custom radiomic sequencer for generating radiomic sequences tailored for lung cancer prediction.  To realize the concept of discovery radiomics for lung cancer prediction in this study, we take inspiration from deep network architectures, which has been shown to be effective for medical image analysis~\cite{roth2015improving,shen2015multi} and introduce a deep convolutional radiomic sequencer that is discovered using a deep convolutional neural network architecture.  The architecture of the deep convolutional radiomic sequencer is shown in Fig.~\ref{cnn_out}.  The proposed radiomic sequencer has 3 convolutional sequencing layers; the number of receptive fields and receptive field size used in each convolutional sequencing layer is specified in Table~\ref{tab_Kernels}.

In between each pair of convolutional sequencing layers there is a max-pooling sequencing layer for improving translational tolerance.  The final output of the deep convolutional radiomic sequencer is a radiomic sequence with 500 abstract imaging-based features.

\vspace{-0.4cm}
\subsubsection{Radiomic Sequencer Discovery}
\vspace{-0.2cm}
To discover the aforementioned deep convolutional radiomic sequencer, we construct a deep convolutional neural network architecture for the radiomic sequencer discovery process, where the radiomic sequencer is directly embedded in the sequencer discovery architecture and learned based on the available CT imaging data, radiologist annotations, and diagnostic results from past patients. In the radiomic sequencer discovery architecture, a fully-connected layer, a rectified linear unit layer, and a loss layer are augmented at the end of the radiomic sequencer to be discovered for the learning process.  The receptive fields in the convolutional sequencing layers of the radiomic sequencer along with the fully-connected layer in the sequencer discovery architecture are learned in a supervised manner based on the input CT imaging data.  This process allows us to learn specialized receptive fields in the custom radiomic sequencer that better characterize the unique tumour traits captured in the CT imaging data in an abstract fashion beyond that  can be captured using pre-defined, hand-engineered features in current radiomics-driven approaches.  A softmax loss function is used in the loss layer of the sequencer discovery process, and  stochastic gradient descent (SGD) is used for optimization in learning the radiomic sequencer.  Furthemore, a rectified linear unit (ReLU) layer is used to introduce non-saturating nonlinearity into the sequencer. We initialize the learning process with a learning rate of 0.001 and complete the learning process in 60 epochs using a batch size of 100. The momentum is fixed to 0.9 with weight decay parameters set to 0.0005 through-out the learning process.
\vspace{-0.4cm}
\subsubsection{Radiomic Sequence Generation}
\vspace{-0.2cm}
After the custom deep convolutional radiomic sequencer has been discovered by the radiomic sequencer discovery process, it can then be used for generating radiomic sequences. As shown in Fig.~\ref{cnn_out}, for a new patient case, we take the CT imaging data pertaining to the patient and feed it into radiomic sequencer to obtain a final custom radiomic sequence with 500 abstract imaging-based features.
\vspace{-0.4cm}
\section{Experimental Results}
\label{exp}
\vspace{-0.3cm}
This section provides details about the experimental setup and obtained results that are used to evaluate the efficacy of the proposed discovery radiomics framework for the purpose of classifying lung lesions as either malignant or benign based on CT imaging data and pathologically-proven diagnostic data.  The first subsection explains the experimental setup including the performance metrics, and the later subsections present the results obtained using the experimental setup for the task of lesion classification.

\vspace{-0.1cm}

\vspace{-0.5cm}
\begin{table}
\centering
 \caption {Comparison of Patient-Level Classification Results Obtained using Discovered Radiomics Sequencers (DRS), CNN Multiple Instance Learning (CNN-MIL)~\cite{shen2016learning} and Deep Autoencoding Radiomic Sequencers (DARS)~\cite{kumar2015lung}. The best results are highlighted in bold.}
 \begin{tabular}{p{3.5cm}p{2.5cm}p{2.5cm}p{2.5cm}}

  \hline\noalign{\smallskip}

 & DRS & CNN-MIL~\cite{shen2016learning} & DARS~\cite{kumar2015lung}\\ [0.5ex]
  \hline\noalign{\smallskip}
  \textbf{Sensitivity (\%)} &  {79.06} & -- & \textbf{83.14} \\
  \textbf{Specificity (\%)} &  \textbf{76.11} & --  & {20.18}\\
  \textbf{Accuracy (\%)} &  \textbf{77.52} & 70.69 & {75.01} \\
  \hline\noalign{\smallskip}
  
\end{tabular}
\\[15pt] \label{tab:class_result2}
\vspace{-0.8cm}
\end{table}
\vspace{-0.2cm}
To create the input data for discovering the radiomics sequencer in the proposed discovery radiomics framework, we first extract the lung CT scans from the LIDC-IDRI dataset~\cite{armato2011lung} for particular patients which have diagnostic data associated with their cases. Based on the diagnostic result, we were able to get lung CT scan images for 97 patients and divided them into two groups: i) malignant, and ii) benign. From each individual scan, we extracted the lesion based on the provided radiologist annotations of the lesion in the particular CT scan image by up to four radiologists for each image which was later verified by pathological data. To mitigate the inter-observability differences, we include all of the annotations provided by each radiologist.  As stated earlier, after the data augmentation process, the enriched dataset contains 42,340 lung lesions composed of 29,956 of malignant lesions and 22,384 benign lesions. To be consistent with the past approaches in this domain, for evaluation purposes, we divided the dataset into two parts: 90\% of the dataset is used for discovering a custom radiomic sequencer, and 10\% of the dataset is used for testing classification performance using the discovered radiomic sequencer. We further divide the 90\% into two parts: 80\% for training and 10\% as a validation set for validating the training process.  A binary decision tree classifier is used for evaluating the efficacy of utilizing the discovered radiomic sequencers for classifying between benign and malignant tumours. Performance is quantitatively determined via sensitivity, specificity and accuracy metrics.
\vspace{-0.3cm}
%
%
%
\vspace{-0.2cm}
\subsection{Results and Analysis}
\vspace{-0.2cm}
The detailed results for the experiment are presented in Table~\ref{tab:class_result2}. Through the 10 fold cross validation using the same setup we obtain an average accuracy of 77.52\% with 79.06\% sensitivity and 76.11\% specificity using the discovered radiomic sequencer. Furthermore, we also present a comparison of the results obtained using the discovered radiomic sequencer (DRS) along with the results obtained using Deep Autoencoding Radiomic Sequencers (DARS)~\cite{kumar2015lung} and the current state-of-the art CNN Multiple Instance Learning approach (CNN-MIL)~\cite{shen2016learning} (see Table~\ref{tab:class_result2}). It is important to note here that while other past works in the area of lung nodule classification have reported higher accuracy for a similar experimental setup and dataset~\cite{shen2015multi,shen2017multi,kuruvilla2014lung}, these methods are based on ratings provided by radiologists as ground truth for malignancy, which can affect the reliability of the experimental results as it introduces additional inter-observer variability from the radiologists. To have a just comparison with the presented approach, we perform comparisons with CNN-MIL and DARS as these methods are based on pathologically-proven lung cancer prediction, which is considered to be the gold standard for ground truth. This results in a much more difficult but more reliable evaluation configuration similar to~\cite{shen2016learning,kumar2015lung}. From Table~\ref{tab:class_result2}, it is evident that proposed discovery radiomic sequencers framework attains much better overall accuracy compared to CNN-MIL~\cite{shen2016learning} and outperforms the DARS~\cite{kumar2015lung} method on both accuracy and sensitivity while producing similar results for sensitivity. These results show that the proposed radiomics discovery framework has strong potential for improving diagnostic accuracy for lung cancer prediction.


\vspace{-0.4cm}
\section{Conclusion}
\vspace{-0.3cm}
In this study, we presented a discovery radiomics framework designed for pathologically proven lung cancer prediction using CT imaging data and diagnostic data. Since the custom radiomic sequencers are dependent on the data on which they are learned, the discovered radiomic sequencers produce highly tailor-made radiomic sequences for the tumour type, in this case lung nodules. Experimental results show that the discovered radiomic sequencers, when used for pathologically proven lung nodule classification, outperform state-of-the art approach when evaluated using the LIDC-IDRI diagnostic dataset. As such, the presented discovery radiomics framework can be a low cost, fast and repeatable way of producing quantitative characterizations of tumour phenotype that has the potential to speed up the screening and diagnosis process while improving consistency and accuracy.

In terms of future work, an area that is worth investigating is on the use of discovery radiomics for risk stratification.  Furthermore, the discovery radiomics framework has the potential to be of high clinical impact in the field of tumour grading and staging analysis.



\vspace{-0.4cm}
\section*{Acknowledgment}
\vspace{-0.2cm}
This research has been supported by the OICR, Canada Research Chairs programs,
NSERC, and the Ministry of R\&I,ON.
\vspace{-0.4cm}
\bibliographystyle{IEEEtran}
\bibliography{IEEEabrv,references}


\end{document}